%% file: main.tex
\documentclass[10pt,twocolumn,letterpaper]{article}

\usepackage[pagenumbers]{cvpr} 

\usepackage{array}
\usepackage{multirow}%
\usepackage{amsmath,amssymb,amsfonts}%
\usepackage{booktabs}%
\usepackage{threeparttable}
\usepackage{amsmath}
\DeclareMathOperator*{\argmax}{argmax}
\usepackage{bm}
\usepackage[table]{xcolor}
\usepackage{arydshln}

\input{preamble}
\definecolor{cvprblue}{rgb}{0.21,0.49,0.74}
\usepackage[pagebackref,breaklinks,colorlinks,allcolors=cvprblue]{hyperref}

\def\paperID{} 
\def\confName{CVPR}
\def\confYear{2026}

\title{Towards Multimodal Domain Generalization with Few Labels}

\author{
    Hongzhao Li$^{1}$ \quad Hao Dong$^{2}$ \quad Hualei Wan$^{1}$ \quad Shupan Li$^{1}$\thanks{Corresponding authors: Shupan Li, Mingliang Xu} \quad Mingliang Xu$^{1}$\footnotemark[1] \\ 
    Muhammad Haris Khan$^{3}$ \\[1ex] 
    $^{1}$Zhengzhou University \qquad $^{2}$ETH Zürich \qquad $^{3}$MBZUAI \\
}
\begin{document}
\maketitle
\input{sec/0_abstract}    
\input{sec/1_intro}

\input{sec/2_related_work}
\input{sec/3_problem_definition}
\input{sec/4_Methodology}
\input{sec/5_Experiments}
\input{sec/6_conclusion}
\section*{Acknowledgments}
This work was supported by the National Key R\&D Program of China (2024YFB3311600) and the Natural Science Foundation of Henan (252300423936).
{
    \small
    \bibliographystyle{ieeenat_fullname}
    \bibliography{main}
}
\input{sec/X_suppl}

\end{document}

%% file: sec/0_abstract.tex
\begin{abstract}
Multimodal models ideally should generalize to unseen domains while remaining data-efficient to reduce annotation costs. To this end, we introduce and study a new problem, Semi-Supervised Multimodal Domain Generalization (SSMDG), which aims to learn robust multimodal models from multi-source data with few labeled samples. We observe that existing approaches fail to address this setting effectively: multimodal domain generalization methods cannot exploit unlabeled data, semi-supervised multimodal learning methods ignore domain shifts, and semi-supervised domain generalization methods are confined to single-modality inputs. To overcome these limitations, we propose a unified framework featuring three key components: Consensus-Driven Consistency Regularization, which obtains reliable pseudo-labels through confident fused-unimodal consensus; Disagreement-Aware Regularization, which effectively utilizes ambiguous non-consensus samples; and Cross-Modal Prototype Alignment, which enforces domain- and modality-invariant representations while promoting robustness under missing modalities via cross-modal translation. We further establish the first SSMDG benchmarks, on which our method consistently outperforms strong baselines in both standard and missing-modality scenarios. Our benchmarks and code are available at \url{https://github.com/lihongzhao99/SSMDG}.
\end{abstract}

%% file: sec/1_intro.tex
\section{Introduction}
\label{sec:intro}
In real-world applications, multimodal models have become increasingly prevalent \cite{yuan2025survey}. However, when deployed in new environments, these models often face domain shifts, where the distribution of test data differs substantially from that of the training data \cite{zhou2022domain,wang2022generalizing,zhou2021domain,dong2025advances}. For instance, an action recognition model trained on high-quality studio recordings may perform poorly on outdoor, user-generated videos due to differences in lighting, background, and audio noise. This performance degradation underscores the need for models that can generalize to unseen domains. To address this challenge, multimodal domain generalization~\cite{dong2023simmmdg,dong2024towards,li2025towards} has emerged, aiming to learn from diverse multimodal source domains to achieve robust cross-domain generalization. Beyond generalization, an ideal learning system should also be data-efficient, capable of learning effectively from a few labeled data to mitigate the high annotation costs of multimodal datasets. This goal aligns closely with semi-supervised learning \cite{van2020survey,sohn2020fixmatch,fini2023semi,yang2022class,rizve2022towards,huang2023contrastive,wang2023freematch,cheng2025cgmatch}, which seeks to exploit abundant unlabeled data alongside a small labeled subset during training.

\begin{figure*}[t]
  \centering
   \includegraphics[width=\linewidth]{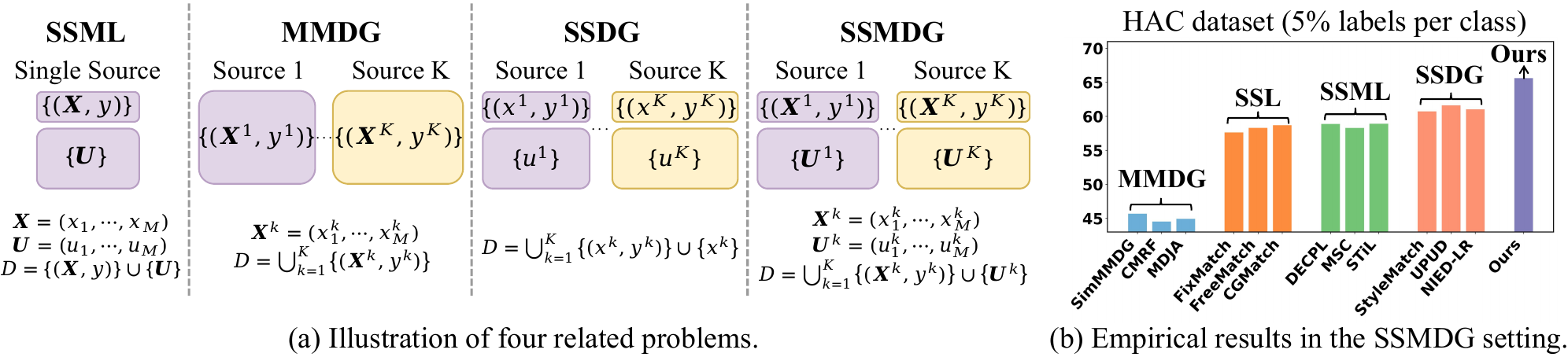}
   \caption{(a) Illustration of four related learning settings: Semi-Supervised Multimodal Learning (SSML), Multimodal Domain Generalization (MMDG), Semi-Supervised Domain Generalization (SSDG), and our proposed Semi-Supervised Multimodal Domain Generalization (SSMDG). SSML overlooks domain shifts, MMDG cannot leverage unlabeled data, and SSDG is restricted to single-modality inputs. (b) Performance comparison in the SSMDG setting, highlighting the limitations of existing paradigms and the effectiveness of our proposed method.}
   \label{fig:1}
\end{figure*}

In this work, we introduce \textbf{Semi-Supervised Multimodal Domain Generalization (SSMDG)}, a novel problem setting that unifies three key aspects of real-world learning: domain generalization, data efficiency, and multimodal learning. SSMDG aims to train models that leverage multimodal data with \textit{limited annotations} from multiple source domains to generalize effectively to unseen target domains.
SSMDG lies at a largely unexplored intersection. As illustrated in Fig.~\ref{fig:1} (a), existing paradigms such as Semi-Supervised Multimodal Learning (SSML) \cite{sun2020tcgm,assefa2024audio,du2025stil}, Multimodal Domain Generalization (MMDG) \cite{dong2023simmmdg,planamente2024relative,li2025towards}, and Semi-Supervised Domain Generalization (SSDG) \cite{zhou2023semi,galappaththige2025domain,lee2025unlocking}, each address only partial aspects of this challenge. Consequently, these methods perform poorly under the SSMDG setting (Fig.~\ref{fig:1} (b)). Their limitations stem from fundamental design gaps: SSML models leverage unlabeled multimodal data but overlook domain shifts; MMDG methods focus on generalization but cannot effectively utilize unlabeled data; SSDG approaches handle domain shifts with limited supervision but are restricted to single-modality inputs, thereby failing to capture cross-modal interactions. These deficiencies highlight the need for a unified framework that can jointly exploit unlabeled multimodal data, learn domain-invariant representations, and operate effectively under sparse supervision.

To address SSMDG, we design a unified framework guided by its two core challenges: (1) identifying reliable pseudo-labels from unlabeled data despite low confidence and inter-modality disagreements, and (2) learning representations that are both modality- and domain-invariant under limited supervision. This combination forms a unique bottleneck distinct from SSML, SSDG, and MMDG. To tackle the first challenge, we introduce a two-part regularization strategy. Consensus-Driven Consistency Regularization enforces consistency only when the fused multimodal prediction and at least one unimodal prediction are both confident and in agreement, ensuring reliable pseudo-labels. Disagreement-Aware Regularization extends this to ambiguous samples with diverging modalities, employing a robust generalized cross-entropy loss to stabilize training while mitigating noise. For the second challenge, we propose Cross-Modal Prototype Alignment to jointly enforce modality and domain invariance. This module leverages class prototypes as shared semantic anchors, aligning multimodal and multi-domain features to construct a stable semantic structure crucial for generalization. It further integrates cross-modal feature translation, enhancing robustness to missing modalities. Our main contributions are summarized as follows:
\begin{itemize}
\item \textbf{New problem.} We introduce Semi-Supervised Multimodal Domain Generalization, unifying the previously disjoint challenges of generalization and data efficiency in real-world multimodal learning.

\item \textbf{Comprehensive benchmark.} We establish the first SSMDG benchmark, covering diverse scenarios for systematic evaluation.

\item \textbf{Novel insights.} We analyze the limitations of existing paradigms and identify the unique challenges specific to the SSMDG setting.

\item \textbf{Effective framework.} We propose a unified framework that effectively addresses the key challenges of SSMDG, enabling robust multimodal learning under domain shifts with few labeled samples.
\end{itemize}

%% file: sec/2_related_work.tex
\section{Related Work} \label{sec:Related Work} 
\noindent\textbf{Semi-Supervised Multimodal Learning} aims to learn from multimodal data with limited annotations \cite{assefa2024audio,sun2020tcgm,chen2025semi,cheng2016semi,hazra2025reflective,li2025semi,du2025stil}. These methods typically exploit cross-modal consistency and complementarity to improve representation learning. Common approaches employ cross-modal contrastive learning or consistency regularization \cite{assefa2024audio} to enforce similar representations or predictions across modalities of the same sample. Other works adopt information-theoretic formulations \cite{sun2020tcgm, chen2025semi} or generative models to enhance multimodal fusion. While effective, these approaches assume a single-domain setting where training and testing data share the same distribution, thus lack mechanisms to handle domain shifts central to generalization.

\begin{figure*}[!t]
  \centering
   \includegraphics[width=\linewidth]{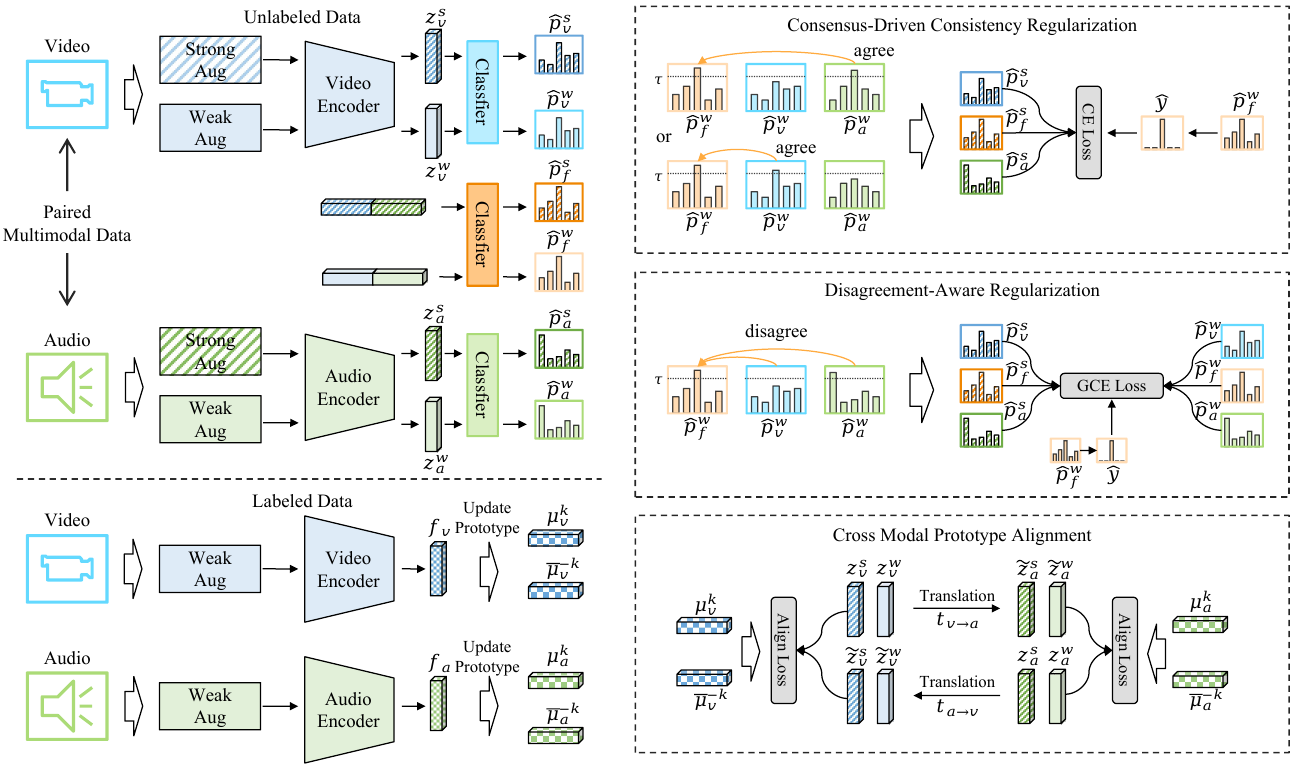}
   \caption{Overview of the proposed framework for SSMDG. The framework jointly leverages labeled and unlabeled multimodal data from multiple source domains through three key components: Consensus-Driven Consistency Regularization (CDCR, \S\ref{sec:CDCR}), Disagreement-Aware Regularization (DAR, \S\ref{sec:DAR}), and Cross Modal Prototype Alignment (CMPA, \S\ref{sec:CMPA}), enabling robust generalization to unseen target domains.}
\label{fig:2}
\end{figure*}

\noindent\textbf{Multimodal Domain Generalization} instead aims to generalize multimodal models to unseen domains using labeled data from multiple source domains \cite{dong2023simmmdg,fan2024cross,huang2025bridging,planamente2024relative,dong2024towards,li2025towards,gungor2025integrating,zhang2025nonpolarized}. These methods focus on learning representations that are invariant to both modality and domain. Typical strategies include extending unimodal domain generalization techniques to multimodal inputs \cite{huang2025bridging}, mapping heterogeneous features into a unified latent space \cite{fan2024cross}, or developing alignment schemes to mitigate domain shifts \cite{planamente2024relative, dong2024towards}. However, MMDG methods operate under a fully supervised paradigm, assuming all source samples are labeled, and therefore cannot leverage the abundant unlabeled data often available in practice.

\noindent\textbf{Semi-Supervised Domain Generalization} extends classic semi-supervised learning frameworks such as FixMatch \cite{sohn2020fixmatch} to the multi-domain setting, learning from limited labeled and abundant unlabeled data \cite{khan2024improving,lee2025unlocking,galappaththige2024towards,galappaththige2025domain}. To enhance domain-invariant feature learning, these methods develop robust pseudo-labeling strategies \cite{lee2025unlocking} or introduce domain-aware mechanisms such as stochastic style matching \cite{zhou2023semi} and domain-guided weight modulation \cite{galappaththige2025domain}. Although effective under sparse supervision, SSDG methods are inherently designed for single-modality inputs, preventing them from exploiting the consensus and complementarity across multiple modalities.

%% file: sec/3_problem_definition.tex
\section{Problem Definition}
\label{sec:Problem Definition}
We study the problem of \textit{Semi-Supervised Multimodal Domain Generalization}. Let $\mathcal{X} = \mathcal{X}_1 \times \cdots \times \mathcal{X}_M$ denote the multimodal input space composed of $M$ distinct modalities, and let $\mathcal{Y}$ represent the shared label space. Each multimodal sample is denoted as $\bm{X} = (\mathbf{x}_1, \dots, \mathbf{x}_M)$, with each modality-specific input $\mathbf{x}_m \in \mathcal{X}_m$. A domain is formally defined by a joint distribution $P(\bm{X}, Y)$ over $\mathcal{X} \times \mathcal{Y}$, with corresponding marginals $P(\bm{X})$ and $P(Y)$.

We assume access to $K$ distinct but related source domains, $\mathcal{S} = \{\mathcal{S}_k\}_{k=1}^K$, each associated with a joint distribution $P^{(k)}(\bm{X}, Y)$. We consider domain shifts that arise solely from variations in the input distribution, i.e., $P^{(k)}(\bm{X}) \neq P^{(k')}(\bm{X})$ for $k \neq k'$, while the label space $\mathcal{Y}$ and its marginal distribution $P(Y)$ remain consistent across all domains. This setup implies that the underlying input-label relationship, captured by the conditional distribution $P(Y|\bm{X})$, is invariant and shared among all domains.

In the semi-supervised setting, the data for each source domain $\mathcal{S}_k$ consists of:
A small labeled subset $\mathcal{D}_k^l = \{(\bm{X}_i^{(k)}, y_i^{(k)})\}_{i=1}^{n_k^l}$, where $(\bm{X}_i^{(k)}, y_i^{(k)})$ are drawn i.i.d. from $P^{(k)}(\bm{X}, Y)$.
A much larger unlabeled subset $\mathcal{D}_k^u = \{\bm{U}_j^{(k)}\}_{j=1}^{n_k^u}$, where $\bm{U}_j^{(k)}$ are drawn from the marginal distribution $P^{(k)}(\bm{X})$. The total size of the unlabeled data significantly outweighs the labeled data, i.e., $n_k^l \ll n_k^u$.
The goal of SSMDG is to leverage all labeled and unlabeled data from the $K$ source domains to learn a single predictive model $f_\theta : \mathcal{X}_1 \times \cdots \times \mathcal{X}_M \to \mathcal{Y}$, parameterized by $\theta$, that generalizes effectively to an unseen target domain $\mathcal{T}$. The target domain is characterized by an unknown distribution $P^{(\mathcal{T})}(\bm{X}, Y)$ that is entirely inaccessible during training. Consistent with our assumptions, $\mathcal{T}$ shares the same invariant conditional distribution $P(Y|\bm{X})$ and label space $P(Y)$ as the source domains, but differs in its input distribution:
\begin{equation}
    P^{(\mathcal{T})}(\bm{X}) \neq P^{(k)}(\bm{X}), \quad \forall k \in \{1, \dots, K\}.
\end{equation}
Model performance is evaluated directly on samples $\bm{X}^* \sim P^{(\mathcal{T})}(\bm{X})$ from the unseen target domain without any adaptation at test time.

%% file: sec/4_Methodology.tex
\section{Methodology}
\label{sec:Methodology}
In this section, we introduce our proposed framework for SSMDG. For clarity, we describe our method using a bimodal case with video ($v$) and audio ($a$), though it naturally generalizes to multiple modalities.
The model comprises modality-specific encoders, unimodal classifiers, and a fusion classifier operating on concatenated features (denoted as $[\cdot ; \cdot]$).
Training proceeds in a mini-batch fashion, sampling from the union of labeled data $\bigcup_k \mathcal{D}_k^l$ and unlabeled data $\bigcup_k \mathcal{D}_k^u$ across $K$ source domains. 

For each labeled sample $(\bm{X}^{(k)},y^{(k)})=((\mathbf{x}_v^{(k)},\mathbf{x}_a^{(k)}), $ $y^{(k)})$ from domain $k$, we apply weak augmentations and encode each modality to obtain $\mathbf{f}_v$ and $\mathbf{f}_a$. From these features, we compute the unimodal predictions $\hat{\mathbf{p}}_v^l$, $\hat{\mathbf{p}}_a^l$ and the fused prediction $\hat{\mathbf{p}}_f^l$ based on $[\mathbf{f}_v; \mathbf{f}_a]$.
For each unlabeled sample $\bm{U}^{(k)} = (\mathbf{u}_v^{(k)}, \mathbf{u}_a^{(k)})$ from domain $k$, we generate two augmented views: a weakly augmented view and a strongly augmented view.
Specifically, weak augmentations ($\mathcal{A}_w$) consist of standard spatial or acoustic transforms like random flips or pitch shifts, while strong augmentations ($\mathcal{A}_s$) employ aggressive perturbations such as RandAugment \cite{cubuk2020randaugment} for video and SpecAugment \cite{park2019specaugment} for audio.
Passing these through the encoders yields weak-view features ($\mathbf{z}_v^w$, $\mathbf{z}_a^w$) and the strong-view features ($\mathbf{z}_v^s$, $\mathbf{z}_a^s$). The corresponding predictions are $\hat{\mathbf{p}}_v^w$, $\hat{\mathbf{p}}_a^w$, $\hat{\mathbf{p}}_f^w$ for the weak view, and $\hat{\mathbf{p}}_v^s$, $\hat{\mathbf{p}}_a^s$, $\hat{\mathbf{p}}_f^s$ for the strong view. All predictions $\hat{\mathbf{p}}$ represent the softmax probabilities over $|\mathcal{Y}|$ classes.

Our framework integrates three complementary components (Fig.~\ref{fig:2}): (1) Consensus-Driven Consistency Regularization, which generates reliable pseudo-labels based on cross-modal agreement; (2) Disagreement-Aware Regularization, which leverages ambiguous non-consensus samples; and (3) Cross Modal Prototype Alignment, which enforces cross-modal and cross-domain feature consistency.

\subsection{Consensus-Driven Consistency Regularization}
\label{sec:CDCR}
In semi-supervised learning, reliable pseudo-labels are essential for effective consistency regularization \cite{yang2022survey}.
Motivated by the idea that joint decisions from multiple perspectives are generally more robust than single-view predictions \cite{du2024improving}, we propose Consensus-Driven Consistency Regularization (CDCR).
This component employs a modality-consensus selection strategy, identifying unlabeled samples that exhibit both high confidence and cross-modal agreement between fused and unimodal predictions.

For each unlabeled sample, we derive a pseudo-label $\hat{y} = \argmax \hat{\mathbf{p}}_f^w$ from the weakly augmented fused prediction. A sample is included in the consensus set $\mathcal{B}^u_{\text{cdcr}}$ if: (1) the fused prediction confidence exceeds a threshold $\tau$, i.e., $\max \hat{\mathbf{p}}_f^w > \tau$; (2) $\hat{y}$ matches at least one unimodal pseudo-label ($\argmax \hat{\mathbf{p}}_v^w$ or $\argmax \hat{\mathbf{p}}_a^w$); and (3) the corresponding unimodal prediction(s) also exceed $\tau$.
For samples that meet these criteria, we enforce prediction consistency across modalities using cross-entropy loss between the pseudo-label and the strong-view predictions:
\begin{equation}
\mathcal{L}_{\text{cdcr}} = \frac{1}{|\mathcal{B}^u_{\text{cdcr}}|} \sum_{\bm{U} \in \mathcal{B}^u_{\text{cdcr}}} \sum_{n \in \{v, a, f\}} \mathcal{H}(\hat{y}, \hat{\mathbf{p}}_n^s),
\end{equation}
where $\mathcal{H}(\cdot, \cdot)$ is the cross-entropy loss. This FixMatch-style \cite{sohn2020fixmatch} regularization encourages the model to maintain consistent predictions across augmented views, leveraging only high-confidence, cross-modally consistent pseudo-labels for supervision.

\subsection{Disagreement-Aware Regularization}
\label{sec:DAR}
While CDCR focuses on reliable samples, it inevitably discards many potentially informative ones. To exploit these, we define a non-consensus set $\mathcal{B}^u_{\text{dar}}$, which includes samples that fail the CDCR consensus criteria but still exhibit high fused prediction confidence ($\max \hat{\mathbf{p}}_f^w > \tau$). For this set, we introduce Disagreement-Aware Regularization (DAR).
DAR leverages the fused pseudo-label as a supervisory signal but replaces the standard cross-entropy with the Generalized Cross-Entropy (GCE) loss \cite{zhang2018generalized}, which is known for its robustness to noisy labels. This makes it particularly suitable for non-consensus samples, where pseudo-labels are more prone to errors.

Formally, for a pseudo-label index $\hat{y}$ and its corresponding probability distribution $\mathbf{p}$, the GCE loss is defined as $\mathcal{L}_{\text{GCE}}(\hat{y}, \mathbf{p}) = (1 - (\mathbf{p})_{\hat{y}}^q) / q$, where $(\mathbf{p})_{\hat{y}}$ is the probability assigned to the class $\hat{y}$, and $q \in (0,1]$ controls robustness. The DAR objective is then formulated as:
\begin{equation}
\mathcal{L}_{\text{dar}} = \frac{1}{|\mathcal{B}^u_{\text{dar}}|} \sum_{\bm{U} \in \mathcal{B}^u_{\text{dar}}} \sum_{j \in \{w, s\}}\sum_{n \in \{v, a, f\}}  \mathcal{L}_{\text{GCE}}(\hat{y}, \hat{\mathbf{p}}_n^j) .
\end{equation}
This objective applies GCE between the weak-fused pseudo-label $\hat{y}$ and all corresponding unimodal and fused predictions from both weak ($j=w$) and strong ($j=s$) views, encouraging stable learning even in the presence of pseudo-label uncertainty.

\begin{table*}
\centering
\caption{\label{tab:5label}Semi-supervised multimodal domain generalization results with 5 labels per class.}
\resizebox{\linewidth}{!}{
\begin{threeparttable}
\begin{tabular}{lccp{2.0cm}p{2.0cm}p{2.0cm}p{1.3cm}p{2.0cm}p{2.0cm}p{2.0cm}c}
\toprule
\multirow{2}{*}{\textbf{Method}}& \multicolumn{2}{c}{\textbf{Modality}} & \multicolumn{4}{c}{\textbf{HAC dataset}}& \multicolumn{4}{c}{\textbf{EPIC-Kitchens dataset}}\\
\cmidrule(lr){2-3} \cmidrule(lr){4-7} \cmidrule(lr){8-11} 
& Video & Audio & A, C $\rightarrow$ H & H, C $\rightarrow$ A & H, A $\rightarrow$ C  & \textit{Mean} & D2, D3 $\rightarrow$ D1 & D1, D3 $\rightarrow$ D2 & D1, D2 $\rightarrow$ D3  & \textit{Mean}\\
\midrule
\multicolumn{11}{c}{\textit{\textbf{Baseline method}}} \\
Source-only & $\checkmark$& $\checkmark$ & 42.15 & 47.20 & 37.82 & 42.39 & 25.31 & 32.12 & 30.95 & 29.46 \\ 
\midrule
\multicolumn{11}{c}{\textit{\textbf{Multimodal domain generalization methods (MMDG)}}} \\
RNA-Net & $\checkmark$& $\checkmark$& 43.82 & 45.63 & 35.44 & 41.63 & 24.14 & 31.87 & 29.32 & 28.44 \\
SimMMDG & $\checkmark$& $\checkmark$& 46.33 & 48.32 & 38.53 & 44.39 & 25.62 & 32.66 & 33.26 & 31.11 \\
MOOSA & $\checkmark$& $\checkmark$& 45.16 & 46.33 & 39.36 & 43.62 & 25.88 & 34.53 & 32.91 & 30.51 \\
CMRF & $\checkmark$& $\checkmark$& 44.10 & 46.88 & 38.40 & 43.13 & 26.31 & 33.49 & 33.62 & 31.21 \\
MDJA & $\checkmark$& $\checkmark$& 47.03 & 47.19 & 38.62 & 44.28 & 25.69 & 34.65 & 34.20 & 31.51 \\
\midrule
\multicolumn{11}{c}{\textit{\textbf{Unimodal semi-supervised learning methods (SSL)}}} \\
FixMatch & $\checkmark$&  & 50.36 & 52.61 & 43.26 & 48.74 & 32.14 & 34.59 & 30.90 & 32.54 \\
FreeMatch & $\checkmark$&  & 51.32 & 51.98 & 42.19 & 48.50 & 32.31 & 35.20 & 32.06 & 33.19 \\
CGMatch & $\checkmark$&  & 51.93 & 52.30 & 43.08 & 49.10 & 33.61 & 34.87 & 31.77 & 33.42 \\
FixMatch &  &  $\checkmark$ & 32.62 & 31.29 & 25.80 & 29.90 & 24.05 & 25.32 & 24.33 & 24.57 \\
FreeMatch &  &  $\checkmark$ & 32.17 & 32.20 & 25.33 & 29.90 & 22.96 & 23.64 & 25.14 & 23.91 \\
CGMatch &  &  $\checkmark$ & 33.05 & 31.59 & 26.42 & 30.35 & 23.87 & 25.23 & 25.88 &  24.99 \\
FixMatch$^{M}$ & $\checkmark$& $\checkmark$ & 59.78 & 57.95 & 45.17 & 54.30 & 34.25 & 37.20 & 36.17 & 35.87  \\
FreeMatch$^{M}$ & $\checkmark$& $\checkmark$ & 58.32 & 58.66 & 44.33 & 53.77 & 33.98 & 36.34 & 36.92 & 35.75 \\
CGMatch$^{M}$ & $\checkmark$& $\checkmark$ & 60.74 & 60.35 & 45.26 & 55.45 & 34.30 & 37.27 & 36.54 & 36.04 \\
\midrule
\multicolumn{11}{c}{\textit{\textbf{Semi-supervised multimodal learning methods (SSML)}}} \\
Co-training & $\checkmark$& $\checkmark$ & 56.31 & 54.68 & 40.32 & 50.44 & 32.64 & 34.67 & 32.06 & 33.12 \\
TCGM & $\checkmark$& $\checkmark$ & 62.22 & 61.09 & 46.62 & 56.64 & 35.71 & 38.33 & 35.52 & 36.52 \\
MSC & $\checkmark$& $\checkmark$ & 62.86 & 62.52 & 45.21 & 56.86 & 34.72 & 36.20 & 34.33 & 35.08 \\
DECPL & $\checkmark$& $\checkmark$ & 63.60 & 61.31 & 45.73 & 56.88 & 34.64 & 38.33 & 35.01 & 35.99 \\
STiL & $\checkmark$& $\checkmark$ & 64.31 & 63.80 & 46.90 & 58.34 & 35.48 & 38.47 & 36.04 & 36.66 \\
\midrule
\multicolumn{11}{c}{\textit{\textbf{Unimodal semi-supervised domain generalization methods (SSDG)}}} \\
StyleMatch & $\checkmark$ &  & 51.82 & 53.50 & 43.99 & 49.77 & 31.69 & 33.96 & 30.69 & 32.11 \\
UPUD & $\checkmark$ &  & 51.96 & 53.46 & 44.01 & 49.81 & 31.44 & 34.62 & 30.94 &  32.33 \\
NIED-LR & $\checkmark$ &  & 52.33 & 54.33 & 43.26 & 49.97 & 32.86 & 35.72 & 31.26 & 33.28 \\
StyleMatch & & $\checkmark$ & 33.20 & 32.80 & 26.23 & 30.74 & 23.31 & 26.02 & 28.03 & 25.79 \\
UPUD & & $\checkmark$ & 32.05 & 33.02 & 26.96 & 30.68 & 23.04 & 25.65 & 26.79 & 25.16 \\
NIED-LR & & $\checkmark$ & 33.28 & 32.49 & 26.52 & 30.76 & 22.88 & 25.32 & 25.82 & 24.67 \\
StyleMatch$^{M}$ & $\checkmark$& $\checkmark$ & 61.69 & 62.93 & 46.02 & 56.88 & 34.48 & 37.33 & 36.04 & 35.95 \\
UPUD$^{M}$ & $\checkmark$& $\checkmark$ & 61.33 & 62.24 & 49.88 & 57.82 & 35.63 & 38.26 & 37.17 & 37.02 \\
NIED-LR$^{M}$ & $\checkmark$& $\checkmark$ & 62.32 & 63.36 & 47.23 & 57.64 & 36.09 & 37.60 & 37.68 & 37.12 \\
\midrule
\multicolumn{11}{c}{\textit{\textbf{Semi-supervised multimodal domain generalization method (SSMDG)}}} \\
Ours  & $\checkmark$& $\checkmark$ & \textbf{65.58} & \textbf{65.34} & \textbf{51.38} & \textbf{60.77}  & \textbf{37.24} & \textbf{41.73} & \textbf{40.86} &  \textbf{39.94} \\
\bottomrule
\end{tabular}
\end{threeparttable}
}
\end{table*}

\subsection{Cross Modal Prototype Alignment}
\label{sec:CMPA}
While CDCR and DAR operate in the prediction space, learning domain-invariant and cross-modally consistent features is crucial for multimodal domain generalization \cite{li2025towards}.
To this end, we propose Cross-Modal Prototype Alignment (CMPA), which aligns features with both intra-domain class prototypes and corresponding prototypes from other modalities and domains.

We maintain running-average class-prototypes $\bm{\mu}_{m,c}^{(k)}$ for each modality $m \in \{v, a\}$, class $c \in \mathcal{Y}$, and domain $k$. These are updated via exponential moving average (EMA) using the set $\mathcal{F}_{m,c}^{(k)}$, defined as the set of labeled features $\mathbf{f}_m$ from the current batch belonging to modality $m$, class $c$, and domain $k$:
\begin{equation}
\bm{\mu}_{m,c}^{(k)} \leftarrow \alpha \bm{\mu}_{m,c}^{(k)} + (1 - \alpha) \cdot \frac{1}{|\mathcal{F}_{m,c}^{(k)}|} \sum_{\mathbf{f}_m \in \mathcal{F}_{m,c}^{(k)}} \mathbf{f}_m,
\end{equation}
where $\alpha$ is the momentum. To facilitate cross-modal alignment and enable robustness to missing modalities, we introduce modality translators $t_{v \to a}$ and $t_{a \to v}$ that map features from one modality to another: $\tilde{\mathbf{z}}_a^j = t_{v \to a}(\mathbf{z}_v^j)$ and $\tilde{\mathbf{z}}_v^j = t_{a \to v}(\mathbf{z}_a^j)$ for $j \in \{w, s\}$. These translators not only support alignment but also handle missing-modality scenarios at inference by synthesizing a missing feature from a present one~\cite{dong2023simmmdg}. For example, if the audio feature is missing, it is synthesized from the video feature via the translator $t_{v \to a}$ (and vice versa).

For all pseudo-labeled unlabeled samples $\bm{U} \in \mathcal{B}^u = \mathcal{B}^u_{\text{cdcr}} \cup \mathcal{B}^u_{\text{dar}}$ with $\hat{y}$, we align both original and translated features to intra-domain prototypes $\bm{\mu}_{m,\hat{y}}^{(k)}$ and cross-domain averages $\bar{\bm{\mu}}_{m,\hat{y}}^{-k} = \frac{1}{K-1} \sum_{k' \neq k} \bm{\mu}_{m,\hat{y}}^{(k')}$ using:
\begin{equation}
\mathcal{L}_{\text{align}}(\mathbf{z}, \bm{\mu}, \bar{\bm{\mu}}) = \|\mathbf{z} - \bm{\mu}\|_2^2 + \|\mathbf{z} - \bar{\bm{\mu}}\|_2^2.
\end{equation}
The total prototype loss is:
\begin{equation}
\begin{split}
&\mathcal{L}_{\text{cmpa}} = \frac{1}{|\mathcal{B}^u|} \sum_{\bm{U} \in \mathcal{B}^u} \sum_{j \in \{w, s\}} \sum_{m \in \{v, a\}} \bigg[ \\
&\quad \mathcal{L}_{\text{align}}(\mathbf{z}_m^j, \bm{\mu}_{m,\hat{y}}^{(k)}, \bar{\bm{\mu}}_{m,\hat{y}}^{-k}) + \quad \mathcal{L}_{\text{align}}(\tilde{\mathbf{z}}_m^j, \bm{\mu}_{m,\hat{y}}^{(k)}, \bar{\bm{\mu}}_{m,\hat{y}}^{-k}) \bigg].
\end{split}
\end{equation}

\subsection{Overall Objective}
For labeled samples in batch $\mathcal{B}^l$, we apply standard supervision:
\begin{equation}
\mathcal{L}_{\text{sup}} = \frac{1}{|\mathcal{B}^l|} \sum_{(\bm{X}, y) \in \mathcal{B}^l} \sum_{n \in \{v, a, f\}} \mathcal{H}(y, \hat{\mathbf{p}}^l_n).
\end{equation}
The overall objective is:
\begin{equation}
\mathcal{L} = \mathcal{L}_{\text{sup}} + \lambda_1 \mathcal{L}_{\text{cdcr}} + \lambda_2 \mathcal{L}_{\text{dar}} + \lambda_3 \mathcal{L}_{\text{cmpa}},
\end{equation}
where $\lambda_1, \lambda_2, \lambda_3$ are hyperparameters. This objective is designed to leverage both labeled and unlabeled data while promoting robust cross-domain and cross-modal generalization.

%% file: sec/5_Experiments.tex
\begin{table}
\centering
\caption{\label{tab:fulllabel}SSMDG results on the HAC dataset with 10, 5\%, and 10\% labels per class. We report the mean accuracy across all domains.}
\resizebox{\linewidth}{!}{
\begin{threeparttable}
\begin{tabular}{lcccccc}
\toprule
\multirow{2}{*}{\textbf{Method}}& \multicolumn{3}{c}{\textbf{Modality}} & \multirow{2}{*}{\textbf{10}}&\multirow{2}{*}{\textbf{5\%}}&\multirow{2}{*}{\textbf{10\%}} \\
\cmidrule(lr){2-4}  
& Video & Audio & Flow &  &  &  \\
\midrule
CGMatch$^{M}$ \cite{cheng2025cgmatch}& $\checkmark$& $\checkmark$ &  & 59.33 & 58.74 & 59.32 \\
STiL \cite{du2025stil}& $\checkmark$& $\checkmark$ &  & 60.66 & 58.96 &  60.21 \\
NIED-LR$^{M}$ \cite{galappaththige2025domain}& $\checkmark$& $\checkmark$ &  & 62.92 & 61.08 & 62.86 \\
Ours  & $\checkmark$& $\checkmark$ &  & \textbf{65.00} & \textbf{65.67} & \textbf{65.90} \\
\midrule
CGMatch$^{M}$ \cite{cheng2025cgmatch}& $\checkmark$&  & $\checkmark$ & 58.05 & 59.62 & 60.33 \\
STiL \cite{du2025stil}& $\checkmark$&  & $\checkmark$ & 60.48  & 60.33 & 61.20 \\
NIED-LR$^{M}$ \cite{galappaththige2025domain}& $\checkmark$&  & $\checkmark$ & 62.23 & 61.62 & 61.82 \\
Ours & $\checkmark$&  & $\checkmark$ & \textbf{64.71} & \textbf{64.03} & \textbf{64.45} \\
\midrule
CGMatch$^{M}$ \cite{cheng2025cgmatch}& & $\checkmark$ & $\checkmark$ & 44.79 & 44.32 & 46.32 \\
STiL \cite{du2025stil}& & $\checkmark$ & $\checkmark$ & 45.90 & 45.77 & 47.08 \\
NIED-LR$^{M}$ \cite{galappaththige2025domain}& & $\checkmark$ & $\checkmark$ & 46.69 & 46.73 & 47.66 \\
Ours & & $\checkmark$ & $\checkmark$ & \textbf{49.88} & \textbf{49.69} & \textbf{50.13} \\
\midrule
CGMatch$^{M}$ \cite{cheng2025cgmatch}& $\checkmark$& $\checkmark$ & $\checkmark$ & 59.81 & 60.21 & 62.57 \\
STiL \cite{du2025stil}& $\checkmark$& $\checkmark$ & $\checkmark$ & 62.83 & 63.18 & 63.11 \\
NIED-LR$^{M}$ \cite{galappaththige2025domain}& $\checkmark$& $\checkmark$ & $\checkmark$ & 63.36 & 64.46 & 63.82 \\
Ours  & $\checkmark$& $\checkmark$ & $\checkmark$ & \textbf{66.95} & \textbf{66.44} & \textbf{66.91} \\
\bottomrule
\end{tabular}
\end{threeparttable}
}
\end{table}

\begin{figure*}
  \centering
   \includegraphics[width=\linewidth]{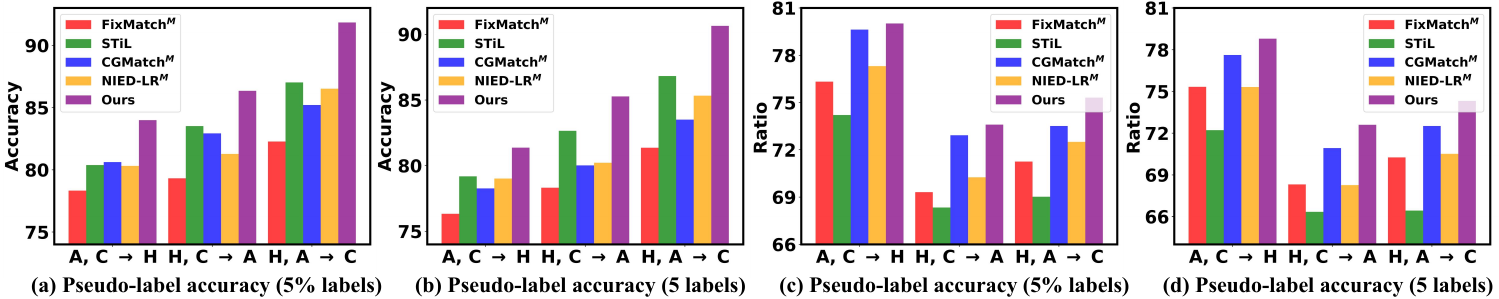}
   \caption{Pseudo-label accuracy and unlabeled data utilization on the HAC benchmark. Our method consistently achieves superior accuracy while maintaining a higher utilization rate compared to competitive baselines.}
   \label{fig:3}
\end{figure*}

\section{Experiments}
\label{sec:Experiments}
\subsection{Settings}
\noindent\textbf{SSMDG Benchmark Datasets:}
We adapt two widely used multimodal domain generalization datasets, EPIC-Kitchens \cite{damen2018scaling} and HAC \cite{dong2023simmmdg}, to benchmark SSMDG methods. EPIC-Kitchens contains eight kitchen actions across three environments (domains D1, D2, D3). HAC comprises seven actions performed by humans, animals, and cartoons (domains H, A, C). 

\noindent\textbf{Evaluation:} 
Following the leave-one-domain-out protocol \cite{zhou2023semi}, two domains are used as source domains for training, and the remaining domain serves as the unseen target for evaluation. We report Top-1 classification accuracy under two SSMDG configurations: (1) 5 or 10 labeled samples per class randomly sampled from each source domain (with the remaining samples being unlabeled); and (2) 5\% or 10\% of samples per class labeled in each source domain. We also assess robustness to missing modalities by simulating unavailable video or audio inputs during testing.

\noindent\textbf{Baselines:}
We compare our approach with state-of-the-art methods across five categories. The Source-only baseline represents the model trained strictly on labeled source domain data without any DG or semi-supervised techniques. The MMDG methods, which rely solely on labeled data, include RNA-NET \cite{planamente2024relative}, SimMMDG \cite{dong2023simmmdg}, MOOSA \cite{dong2024towards}, CMRF \cite{fan2024cross}, and MDJA \cite{li2025towards}. The SSL group includes FixMatch \cite{sohn2020fixmatch}, FreeMatch \cite{wang2023freematch}, and CGMatch \cite{cheng2025cgmatch}. The SSML methods consist of Co-training \cite{blum1998combining}, TCGM \cite{sun2020tcgm}, MSC \cite{chen2025semi}, DECPL \cite{assefa2024audio} and STiL \cite{du2025stil}. Finally, the SSDG group comprises StyleMatch \cite{zhou2023semi}, UPUD \cite{lee2025unlocking}, and NIED-LR \cite{galappaththige2025domain}. SSL and SSDG methods are evaluated in both their original unimodal versions and multimodal extensions (denoted as $^M$) for fairness.

\begin{table*}
\centering
\caption{\label{tab:missing}SSMDG with missing modalities. $\times$ means the modality is available during training, but is missing in test time.}
\resizebox{\linewidth}{!}{
\begin{threeparttable}
\begin{tabular}{lccp{2.0cm}p{2.0cm}p{2.0cm}p{1.3cm}p{2.0cm}p{2.0cm}p{2.0cm}c}
\toprule
\multirow{2}{*}{\textbf{Method}}& \multicolumn{2}{c}{\textbf{Modality}} & \multicolumn{4}{c}{\textbf{HAC dataset}}& \multicolumn{4}{c}{\textbf{EPIC-Kitchens dataset}}\\
\cmidrule(lr){2-3} \cmidrule(lr){4-7} \cmidrule(lr){8-11} 
& Video & Audio & A, C $\rightarrow$ H & H, C $\rightarrow$ A & H, A $\rightarrow$ C  & \textit{Mean} & D2, D3 $\rightarrow$ D1 & D1, D3 $\rightarrow$ D2 & D1, D2 $\rightarrow$ D3  & \textit{Mean}\\
\midrule
\multicolumn{11}{c}{\textit{\textbf{5\% labels per class}}} \\
\midrule
Audio-only & & $\checkmark$ & 32.82 & 31.06 & 24.36 & 29.41 & 25.62 & 25.01 & 26.62 & 25.75 \\
Zero-filling & $\times$ & $\checkmark$ & 30.86 & 30.29 & 23.66 & 28.27 & 24.91 & 25.04 & 26.31 & 25.42 \\
Translation & $\times$ & $\checkmark$ & \textbf{37.88} & \textbf{38.09} & \textbf{31.62} & \textbf{35.86} & \textbf{31.06} & \textbf{30.62 } & \textbf{31.58} & \textbf{31.09} \\
\hdashline
Video-only & $\checkmark$ &  & 51.62 & 52.98 & 48.62 & 51.07 & 35.92 & 35.60 & 34.69 & 35.40 \\
Zero-filling & $\checkmark$& $\times$ & 53.79 & 54.26 & 49.30 & 52.45 & 36.78 & 36.92 & 35.41 & 36.37 \\
Translation & $\checkmark$& $\times$ & \textbf{55.79} & \textbf{57.62} & \textbf{51.68} & \textbf{55.03} & \textbf{39.27} & \textbf{38.64} & \textbf{39.07} & \textbf{38.99} \\
\midrule
\multicolumn{11}{c}{\textit{\textbf{5 labels per class}}} \\
\midrule
Audio-only & & $\checkmark$ & 33.54 & 31.92 & 24.84 & 30.10 & 24.63 & 25.60 & 24.18 & 24.80 \\
Zero-filling & $\times$ & $\checkmark$ & 32.71 & 30.68 & 23.05 & 28.81 & 25.85 & 26.17 & 25.47 & 25.83 \\
Translation & $\times$ & $\checkmark$ & \textbf{39.01} & \textbf{37.55} & \textbf{29.54} & \textbf{35.37} & \textbf{28.88} & \textbf{29.04} & \textbf{27.96} & \textbf{28.63} \\
\hdashline
Video-only & $\checkmark$ &  & 51.08 & 52.66 & 43.86 & 49.20 & 35.37 & 32.69 & 33.33 & 33.80 \\
Zero-filling & $\checkmark$& $\times$ & 51.06 & 52.66 & 43.77 & 49.16 & 36.74 & 32.81 & 33.26 & 34.27 \\
Translation & $\checkmark$& $\times$ & \textbf{53.44} & \textbf{55.82} & \textbf{45.69} & \textbf{51.65} & \textbf{38.15} & \textbf{36.82} & \textbf{37.83} & \textbf{37.60} \\
\bottomrule
\end{tabular}
\end{threeparttable}
}
\end{table*}

\begin{table*}
\centering
\caption{\label{tab:abl} Ablations of each proposed module under 5 labels per class.}
\resizebox{\linewidth}{!}{
\begin{threeparttable}
\begin{tabular}{cccp{2.0cm}p{2.0cm}p{2.0cm}p{1.3cm}p{2.0cm}p{2.0cm}p{2.0cm}c}
\toprule
\multirow{2}{*}{\textbf{CDCR}}& \multirow{2}{*}{\textbf{DAR}} & \multirow{2}{*}{\textbf{CMPA}} & \multicolumn{4}{c}{\textbf{HAC dataset}}& \multicolumn{4}{c}{\textbf{EPIC-Kitchens dataset}}\\
 \cmidrule(lr){4-7} \cmidrule(lr){8-11} 
&  &  & A, C $\rightarrow$ H & H, C $\rightarrow$ A & H, A $\rightarrow$ C  & \textit{Mean} & D2, D3 $\rightarrow$ D1 & D1, D3 $\rightarrow$ D2 & D1, D2 $\rightarrow$ D3  & \textit{Mean}\\
\midrule
 &  &  & 48.62 & 50.28 & 37.26 & 45.39 & 29.06 & 32.74 & 33.92 & 31.91 \\
$\checkmark$ &  &  & 59.92 & 58.67 & 45.69 &  54.76 & 34.72 & 37.69 & 37.02 & 36.48 \\
& $\checkmark$ &  & 60.69 & 57.95 & 46.03 &  54.89 & 34.63 & 37.62 & 36.97 & 36.41 \\
$\checkmark$ & $\checkmark$ &  & 63.37 & 64.26 & 48.28 &  58.64 & 36.02 & 38.63 & 38.91 & 37.85 \\
$\checkmark$ &  & $\checkmark$ & 64.37 & 63.69 & 47.58 & 58.55  & 36.72 & 37.16 & 37.09 & 36.99 \\
 & $\checkmark$ & $\checkmark$ & 63.76 & 62.73 & 48.09 &  58.19 & 35.86 & 37.23 & 36.90 & 36.66 \\
$\checkmark$ & $\checkmark$ & $\checkmark$ & \textbf{65.58} & \textbf{65.34} & \textbf{51.38} &  \textbf{60.77} & \textbf{37.24} & \textbf{41.73} & \textbf{40.86} & \textbf{39.94} \\
\bottomrule
\end{tabular}
\end{threeparttable}
}
\end{table*}

\subsection{Main Results}
\noindent\textbf{SSMDG Performance:} 
We comprehensively benchmark our framework against four related paradigms: MMDG, SSL, SSDG, and SSML. Table \ref{tab:5label} reports results for the challenging 5-label-per-class setting, while Table \ref{tab:fulllabel} summarizes mean performance against top baselines (CGMatch$^M$ \cite{cheng2025cgmatch}, STiL \cite{du2025stil}, NIED-LR$^M$ \cite{galappaththige2025domain}) across various label configurations (10, 5\%, 10\%) and modality combinations (2- and 3-modality). Our framework establishes a new state-of-the-art across all configurations, highlighting the limitations of existing methods under this composite SSMDG challenge.

We observe that each paradigm exhibits a key limitation under the SSMDG setting: (1) MMDG methods (e.g., MDJA \cite{li2025towards}) struggle under extreme label scarcity (Table \ref{tab:5label}), demonstrating their reliance on fully labeled data and inability to leverage unlabeled samples. (2) SSL methods (e.g., CGMatch$^M$ \cite{cheng2025cgmatch}), while using unlabeled data, are not designed for domain generalization; their simple multimodal extensions consequently fail to bridge the domain gap. (3) SSML methods (e.g., STiL \cite{du2025stil}), which handle sparse multimodal labels but lack domain generalization capabilities, resulting in poor performance on unseen target domains. (4) SSDG methods (e.g., NIED-LR$^M$ \cite{galappaththige2025domain}) are designed for unimodal shifts, but their simple multimodal adaptations ($^M$) fail to address the coupled challenges of inter-modal semantics and multimodal domain shifts. 

In contrast, our framework is explicitly designed to simultaneously address sparse labels, domain shifts, and multimodal interactions.
In the 5-label-per-class setting (Table~\ref{tab:5label}), we achieve mean accuracies of 60.77\% on HAC and 39.94\% on EPIC-Kitchens, substantially outperforming all baselines.
Table \ref{tab:fulllabel} further confirms that our superiority is consistent across label regimes and modality combinations.
Importantly, our method scales effectively to three-modality inputs (Video-Audio-Flow), demonstrating its robustness and suitability for complex SSMDG tasks.

\noindent\textbf{Pseudo-Label Quality and Unlabeled Data Utilization:} 
We analyze pseudo-label quality and unlabeled data utilization on the HAC benchmark under the 5-label and 5\%-label settings, comparing our framework with competitive baselines including FixMatch$^M$ \cite{sohn2020fixmatch}, STiL \cite{du2025stil}, StyleMatch$^M$ \cite{zhou2023semi}, and NIED-LR$^M$ \cite{galappaththige2025domain}. As illustrated in Fig.~\ref{fig:3}, our method achieves higher pseudo-label accuracy while maintaining a greater data utilization rate. These findings confirm that our framework effectively extracts reliable supervision from abundant unlabeled multimodal data.

\noindent\textbf{Missing-Modality SSMDG:}
In practical scenarios, some modalities may be unavailable at test time. Table \ref{tab:missing} evaluates robustness to missing video or audio data, comparing unimodal baselines, naive zero-filling, and our cross-modal translation strategy.
Our approach leverages trained translators ($t_{v \to a}, t_{a \to v}$) to generate semantically consistent features for the missing modality, mitigating performance degradation. In contrast, naive zero-filling often underperforms in the shared feature space. For instance, in the 5\%-label setting with a missing video modality, our translation strategy outperforms zero-filling by 7.59\% and 5.67\% on all benchmarks. These results demonstrate that our method can effectively reconstruct missing semantic information while preserving multimodal generalization.

\subsection{Ablation Studies}
\noindent\textbf{Efficacy of Key Model Components:}
Table \ref{tab:abl} evaluates the contributions of CDCR, DAR, and CMPA under the 5-label setting. The weakly supervised baseline (Row 1) highlights that generalization is severely constrained under label scarcity. Incorporating CDCR or DAR individually yields notable improvements, while their combination (Row 4) further boosts performance by jointly leveraging reliable consensus and ambiguous non-consensus samples. CMPA consistently improves all configurations by enforcing feature-space invariance, complementing the prediction-level regularizers. The full framework (Row 7) achieves the best performance, improving HAC accuracy by 15.38\% over the baseline.

\noindent\textbf{Ablation Study on CDCR:} 
We compare alternative consensus strategies in Table \ref{tab:idv-ablation}. Mean-CDCR, similar to the STiL \cite{du2025stil} method, averages predictions from all modalities meeting the threshold, Any2-CDCR requires agreement from any two views, and Strict-CDCR mandates consensus across all views. Our Full-CDCR, which anchors on the fused prediction supported by at least one modality, achieves the best results, outperforming Mean-CDCR by 2.51\% (HAC) and 2.33\% (EPIC-Kitchens), and also surpasses the stricter (Strict-CDCR) and more relaxed (Any2-CDCR) variants, validating our selection design.

\begin{table}
\centering
\caption{Ablation study results demonstrating the effectiveness of individual components under 5 labels per class.}
\resizebox{0.85\linewidth}{!}{
\begin{tabular}{c|cc}
\hline
Method & HAC & EPIC-Kitchens \\
\hline
Mean-CDCR & 58.26 & 37.61 \\
Any2-CDCR & 59.34 & 38.25 \\
Strict-CDCR & 59.16 & 37.32 \\
\rowcolor{gray!20} Full-CDCR & \textbf{60.77} & \textbf{39.94} \\
\hline
CE-DAR & 58.32 & 37.92 \\
Weak-only DAR & 60.42 & 38.97 \\
Strong-only DAR & 60.73 & 39.62 \\
\rowcolor{gray!20} Full-DAR & \textbf{60.77} & \textbf{39.94} \\
\hline
Intra-Domain CMPA & 58.34 & 38.45 \\
Intra-Modal CMPA & 60.75 & 39.64 \\
Weak-only CMPA & 60.22 & 38.64 \\ 
Strong-only CMPA & 59.38 & 39.25 \\
\rowcolor{gray!20} Full-CMPA & \textbf{60.77} & \textbf{39.94} \\
\hline
\end{tabular}}
\label{tab:idv-ablation}
\end{table}

\begin{figure}
  \centering
   \includegraphics[width=\linewidth]{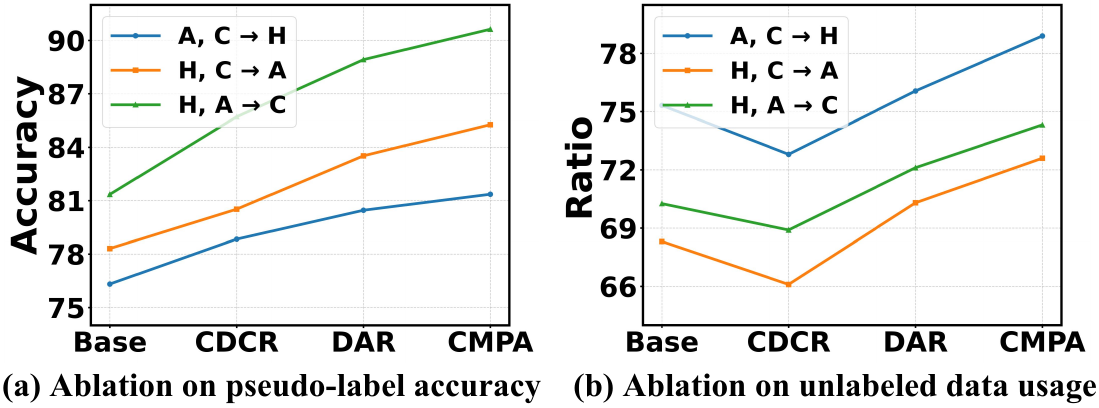}
   \caption{Ablation analysis of different modules on pseudo-label accuracy and unlabeled data utilization.}
   \label{fig:4}
\end{figure}

\noindent\textbf{Ablation Study on DAR:} 
Table \ref{tab:idv-ablation} evaluates DAR design choices. Replacing the GCE loss with standard cross-entropy (CE-DAR) reduces performance (2.45\% on HAC, 2.02\% on EPIC-Kitchens), highlighting the need for a noise-robust loss for ambiguous non-consensus samples. Applying DAR to only weak (Weak-only) or strong (Strong-only) views is suboptimal, whereas Full-DAR, enforcing consistency across both views, achieves the highest accuracy. This demonstrates the advantage of combined regularization.

\noindent\textbf{Ablation Study on CMPA:} 
We investigate CMPA design choices in Table \ref{tab:idv-ablation}. Removing cross-domain alignment (Intra-Domain CMPA) leads to a significant drop (2.43\% on HAC, 1.49\% on EPIC-Kitchens), highlighting the importance of cross-domain feature invariance. Omitting cross-modal translation (Intra-Modal CMPA) also degrades performance, confirming the benefit of inter-modal semantic consistency. Finally, aligning both weak and strong views (Full-CMPA) outperforms single-view variants, validating the joint alignment strategy.

\noindent\textbf{Impact on Pseudo-Labeling:} We assess each module’s effect on pseudo-label quality and unlabeled data utilization on HAC (5-label setting), using FixMatch \cite{sohn2020fixmatch} as the baseline. Fig.~\ref{fig:4} (a) shows that each component progressively improves pseudo-label accuracy. Fig.~\ref{fig:4} (b) illustrates that CDCR initially reduces utilization due to its strict consensus filtering. However, the subsequent addition of DAR and CMPA significantly increases utilization: CMPA stabilizes feature representations, generating more consistent predictions that satisfy confidence and consensus thresholds.

%% file: sec/6_conclusion.tex
\section{Conclusion} 
In this work, we introduced Semi-Supervised Multimodal Domain Generalization (SSMDG), a novel problem that unifies the critical challenges of cross-domain generalization and data-efficient multimodal learning. We analyzed the unique difficulties inherent to SSMDG and highlighted the limitations of existing SSML, MMDG, and SSDG paradigms in addressing them. To bridge this gap, we established the first comprehensive SSMDG benchmarks, including evaluations under missing-modality scenarios, and proposed a new framework that effectively leverages unlabeled multimodal data while enforcing cross-domain and cross-modal invariance. By navigating the delicate balance between cross-modal consensus and informative disagreement, this work provides a practical pathway for deploying multimodal models that remain resilient even when data is scarce and environments are unpredictable. Our method sets a strong performance baseline, and we hope that our problem formulation, benchmarks, and framework will stimulate further research in this practical and impactful direction.

%% file: sec/X_suppl.tex
\clearpage
\setcounter{page}{1}
\maketitlesupplementary

\section{Training Details}
\label{sec:supp_training}
\noindent\textbf{Data Augmentation.}
To make effective use of unlabeled data through consistency regularization, we design a modality-specific augmentation pipeline that clearly separates weak and strong perturbations for video, audio, and optical flow.
\begin{itemize}
    \item Video: Weak augmentation includes random horizontal flips and spatial translations. Strong augmentation applies RandAugment~\cite{cubuk2020randaugment} followed by Cutout~\cite{devries2017improved} to introduce more substantial appearance variations.
    \item Audio: Weak augmentation consists of random gain changes and pitch shifts. Strong augmentation adopts SpecAugment~\cite{park2019specaugment} (frequency and time masking) together with additive noise to simulate acoustic variations and corruptions common in diverse domains.
    \item Optical Flow: We follow a strategy analogous to the video pipeline. Weak augmentation uses random flips and translations, while strong augmentation applies Cutout~\cite{devries2017improved} and noise injection to the flow fields to prevent overfitting to specific motion artifacts.
\end{itemize}

\noindent\textbf{Network Architectures.}
Our framework is implemented using the MMAction2 \cite{contributors2020openmmlab} toolkit. The specific encoders for each modality are as follows:
\begin{itemize}
    \item Video Encoder: We employ a SlowFast network \cite{feichtenhofer2019slowfast} pre-trained on Kinetics-400 \cite{kay2017kinetics}. The output feature dimension is $d_v = 2304$.
    \item Audio Encoder: We utilize a ResNet-18 \cite{he2016deep} architecture, pre-trained on VGGSound \cite{chen2020vggsound}. The output feature dimension is $d_a = 512$.
    \item Optical Flow Encoder: We use a SlowFast network \cite{feichtenhofer2019slowfast} configured with a slow-only pathway, initialized with Kinetics-400 \cite{kay2017kinetics} weights. The feature dimension is $d_f = 2048$.
    \item Cross-Modal Translators: To enable the CMPA module, the translators ($t_{v \to a}, t_{a \to v}$) are modeled as two-layer Multi-Layer Perceptrons (MLP) with 2048 hidden units and ReLU activation, projecting features between modality-specific subspaces.
\end{itemize}

\begin{figure}
  \centering
   \includegraphics[width=0.93\linewidth]{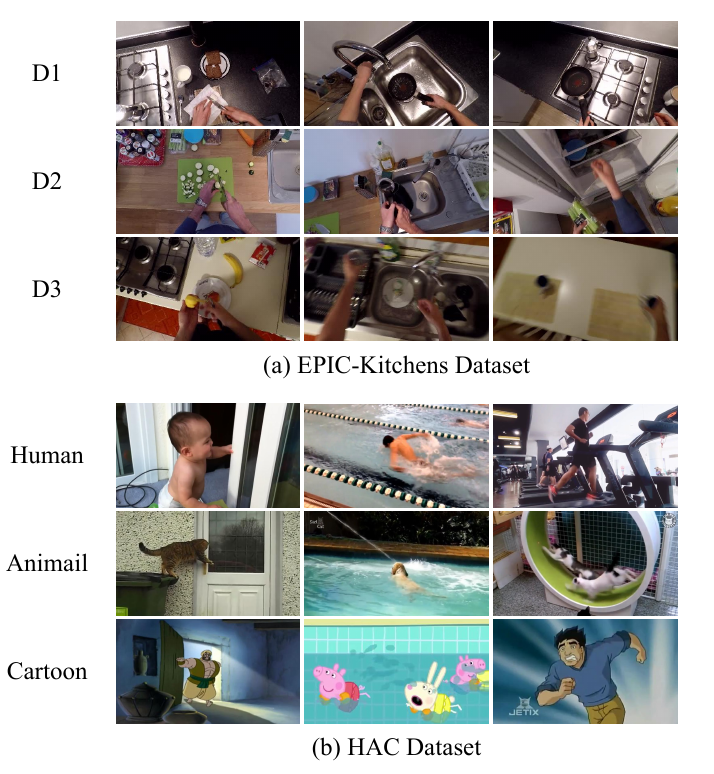}
   \vspace{-0.3cm}
   \caption{Visualization of domain shifts in the experimental benchmarks. (a) Example frames from the EPIC-Kitchens dataset across three environments (D1, D2, D3), highlighting variations in viewpoint and illumination. (b) Samples from the HAC dataset (Human, Animal, Cartoon), illustrating stylistic differences across domains.}
   \label{fig:s1}
\end{figure}

\noindent\textbf{Optimization and Hyperparameters.}
We optimize the entire framework using the AdamW optimizer \cite{loshchilov2017decoupled} with a base learning rate of $1 \times 10^{-4}$, a weight decay of $1 \times 10^{-3}$, and a batch size of 32. The loss balancing hyperparameters are set to $\lambda_1 = 1.0$, $\lambda_2 = 0.1$, and $\lambda_3 = 0.1$. The confidence threshold for pseudo-labeling is set to $\tau = 0.95$. The robustness parameter for the Generalized Cross-Entropy loss in the DAR module is set to $q = 0.7$. All experiments are conducted on two NVIDIA RTX 4090 GPUs.

\section{Dataset Details}
\label{sec:supp_datasets}
\noindent\textbf{EPIC-Kitchens.}
We utilize the domain adaptation subset of EPIC-Kitchens \cite{damen2018scaling}, a large-scale egocentric video dataset. Following standard MMDG protocols \cite{dong2023simmmdg}, we organize the data into three domains (D1, D2, D3), representing three distinct kitchen environments, as shown in Fig.~\ref{fig:s1}~(a). These domains differ in terms of lighting, spatial layout, and background clutter. The dataset contains 8 overlapping action classes: `put', `take', `open', `close', `wash', `cut', `mix', and `pour'. This setup evaluates the model's ability to generalize egocentric action recognition across environmental changes.

\noindent\textbf{HAC.}
The Human-Animal-Cartoon (HAC) dataset \cite{dong2023simmmdg} is employed to evaluate domain generalization under stylistic shifts. The dataset comprises three visually disjoint domains: Human (H), Animal (A), and Cartoon (C), as illustrated in Fig.~\ref{fig:s1}~(b). This setup challenges the model to learn action semantics that are invariant across appearance discrepancies and kinematic variations. The dataset consists of 7 common action classes: `sleep', `watch TV', `eat', `drink', `swim', `run', and `open door'.

\section{Implementation Details of Baselines}
\label{sec:baseline_details}
To ensure a fair comparison, all baseline methods are re-implemented using the MMAction2 \cite{contributors2020openmmlab} codebase. We strictly maintain the same backbone architectures (SlowFast \cite{feichtenhofer2019slowfast} for video, ResNet-18 \cite{he2016deep} for audio). Consistent with our proposed framework, all methods are optimized using AdamW \cite{loshchilov2017decoupled} with a base learning rate of $1 \times 10^{-4}$ and a batch size of 32.

\noindent\textbf{MMDG Baselines.} 
These methods use only the labeled subset of source domains to learn domain-agnostic features.
\begin{itemize}
    \item \textbf{RNA-NET \cite{planamente2024relative}:} We incorporate the Relative Norm Alignment (RNA) loss as a regularization term alongside standard cross-entropy. By minimizing the discrepancy between feature norms of video and audio modalities relative to their class prototypes, RNA-NET aligns the modality distributions to improve generalization.
    \item \textbf{SimMMDG \cite{dong2023simmmdg}:} Following the official implementation, we employ a decomposition head to disentangle features into modality-specific and modality-shared components. We apply supervised contrastive learning to the shared features to enforce semantic consistency, while applying distance constraints to the specific features to maintain modality distinctiveness.
    \item \textbf{MOOSA \cite{dong2024towards}:} We implement two self-supervised pre-tasks: Masked Cross-modal Translation and Multimodal Jigsaw Puzzles. These tasks are optimized jointly with the primary classification objective, encouraging the model to learn robust, domain-invariant representations.
    \item \textbf{CMRF \cite{fan2024cross}:} We implement Cross-Modal Representation Flattening by applying convex combination interpolation on multimodal feature representations. A distillation loss constrains the unimodal networks, enforcing a flatter loss landscape to mitigate the competitive imbalance between modalities.
    \item \textbf{MDJA \cite{li2025towards}:} We implement Modality-Domain Joint Adversarial training by attaching domain discriminators to both unimodal and fused feature extractors. The model utilizes a Gradient Reversal Layer (GRL) to learn features that are discriminative for class boundaries but invariant to domain shifts.
\end{itemize}

\noindent\textbf{SSL Baselines (Extended to Multimodal).} 
We extend original unimodal SSL methods to the multimodal setting (denoted as $^M$) by performing late fusion on features before applying the respective semi-supervised strategies.
\begin{itemize}
    \item \textbf{FixMatch$^M$ \cite{sohn2020fixmatch}:} We apply weak and strong augmentations to all input modalities. A pseudo-label is generated from the fused prediction of the weakly augmented view. This pseudo-label supervises the fused prediction of the strongly augmented view, provided the prediction confidence exceeds a fixed threshold ($\tau=0.95$).
    \item \textbf{FreeMatch$^M$ \cite{wang2023freematch}:} We replace the fixed threshold of FixMatch with Self-Adaptive Thresholding (SAT). The threshold is dynamically adjusted based on the exponential moving average (EMA) of the model's confidence on unlabeled data, improving the utilization of hard classes and noisy domains.
    \item \textbf{CGMatch$^M$ \cite{cheng2025cgmatch}:} CGMatch utilizes the Count-Gap (CG) metric to filter high-value unlabeled samples. We compute the CG score using multimodal predictions and apply the Filtering-by-Dynamic-Selection rule to retain reliable pseudo-labeled samples, strictly following the authors' open-sourced training recipes.
\end{itemize}

\noindent\textbf{SSML Baselines.} 
These methods are inherently designed to handle multimodal data dynamics under label scarcity.
\begin{itemize}
    \item \textbf{Co-training \cite{blum1998combining}:} Treating video and audio as distinct views, we train two independent classifiers on the labeled data. In an iterative process, high-confidence predictions from the video classifier generate pseudo-labels for the Audio classifier, and vice-versa, leveraging the independence of the modalities.
    \item \textbf{TCGM \cite{sun2020tcgm}:} We implement Total Correlation Gain Maximization by adding an information-theoretic loss term to the objective. This maximizes the mutual information (Total Correlation) between video and audio modalities, utilizing unlabeled data to uncover latent semantic structures shared across views.
    \item \textbf{MSC \cite{chen2025semi}:} We implement the Strategic Complementarity learning mechanism. The model dynamically weights the contribution of each modality's prediction based on estimated reliability and consistency, allowing the stronger modality to rectify the weaker one during semi-supervised training.
    \item \textbf{DECPL \cite{assefa2024audio}:} Adopting the Audio-Visual Contrastive learning paradigm, we apply a contrastive loss between audio and video representations alongside consistency regularization. This ensures semantic alignment across modalities is maintained even in the absence of ground-truth labels.
    \item \textbf{STiL \cite{du2025stil}:} Originally designed for tabular-image data, we adapt STiL for the Video-Audio setting. We decompose multimodal representations into shared and specific subspaces via disentangled contrastive consistency and use a consensus-based pseudo-labeling strategy to filter unlabeled samples, refining labels via prototype-guided smoothing.
\end{itemize}

\noindent\textbf{SSDG Baselines (Extended to Multimodal).} 
We adapt methods designed for semi-supervised domain generalization to handle multimodal inputs, focusing on cross-domain robustness.
\begin{itemize}
    \item \textbf{StyleMatch$^M$ \cite{zhou2023semi}:} We extend the stochastic style consistency strategy to the multimodal. We apply stochastic style transfer augmentation to video and audio inputs to simulate domain shifts. A consistency loss is then enforced between the predictions of the original and perturbed multimodal inputs.
    \item \textbf{UPUD$^M$ \cite{lee2025unlocking}:} We implement the Unlabeled Proxy-based Contrastive (UPC) and Surrogate Class (SC) components to better exploit low-confidence unlabeled samples. We extend the method to the multimodal scenario by applying the UPC and SC objectives on the fused video-audio embeddings.
    \item \textbf{NIED-LR$^M$ \cite{galappaththige2025domain}:} We integrate the Domain-Guided Weight Modulation (DGWM) module into a FixMatch backbone. Using available source domain labels, DGWM modulates classifier weights to be domain-aware, thereby enhancing cross-domain generalization capabilities.
\end{itemize}